\documentclass[sigconf,screen]{acmart}

\renewcommand\footnotetextcopyrightpermission[1]{} 
\usepackage{graphicx}
\usepackage{subcaption}
\AtBeginDocument{%
  }

\setcopyright{acmlicensed}
\copyrightyear{2025}
\acmYear{2025}
\acmDOI{XXXXXXX.XXXXXXX}
\acmConference[MM '25]{Proceedings of the 33rd ACM International Conference on Multimedia}{October 27--31, 2025}{Dublin, Ireland}
\acmISBN{978-1-4503-XXXX-X/2018/06}

\begin{document}

\title{GPS: Distilling Compact Memories via Grid-based Patch Sampling for Efficient Online Class-Incremental Learning}


\author{Mingchuan Ma}
\authornote{First author.}
\affiliation{%
  \institution{Sichuan University}
  \city{Chengdu}
  \state{Sichuan}
  \country{China}
}
\email{2022141460145@stu.scu.edu.cn}

\author{Yuhao Zhou}
\authornote{Corresponding author.}
\affiliation{%
  \institution{Sichuan University}
  \city{Chengdu}
  \state{Sichuan}
  \country{China}
}
\email{sooptq@gmail.com}

\author{Jindi Lv}
\affiliation{%
  \institution{Sichuan University}
  \city{Chengdu}
  \state{Sichuan}
  \country{China}
}
\email{lvjindi@stu.scu.edu.cn}

\author{Yuxin Tian}
\affiliation{%
  \institution{Sichuan University}
  \city{Chengdu}
  \state{Sichuan}
  \country{China}
}
\email{cs.yuxintian@outlook.com}

\author{Dan Si}
\affiliation{%
  \institution{Sichuan University}
  \city{Chengdu}
  \state{Sichuan}
  \country{China}
}
\email{sidan_starry@stu.scu.edu.cn}

\author{Shujian Li}
\affiliation{%
  \institution{Sichuan University}
  \city{Chengdu}
  \state{Sichuan}
  \country{China}
}

\author{Qing Ye}
\affiliation{%
  \institution{Sichuan University}
  \city{Chengdu}
  \state{Sichuan}
  \country{China}
}
\email{yeqing@scu.edu.cn}

\author{Jiancheng Lv}
\affiliation{%
  \institution{Sichuan University}
  \city{Chengdu}
  \state{Sichuan}
  \country{China}
}
\email{lvjiancheng@scu.edu.cn}

\renewcommand{\shortauthors}{Ma et al.}

\begin{abstract}
  Online class-incremental learning aims to enable models to continuously adapt to new classes with limited access to past data, while mitigating catastrophic forgetting.
  Conventional replay-based methods address this by maintaining a small memory buffer of previous samples, achieving competitive performance.
  To achieve effective replay under constrained storage, recent approaches leverage distilled data to enhance the informativeness of memory.
  However, such approaches often involve significant computational overhead due to the
  usage of bi-level optimization.
  To address these limitations, we introduce \textbf{G}rid-based \textbf{P}atch \textbf{S}ampling (GPS), a lightweight and effective strategy for distilling informative memory samples without relying on a trainable model. 
  GPS generates informative samples by sampling a subset of pixels from the original image, 
  yielding compact low-resolution representations that preserve both semantic content and structural information. 
  During replay, these representations are reassembled to support training and evaluation. 
  Experiments on extensive benchmarks demonstrate that GRS can be seamlessly integrated into existing replay frameworks, 
  leading to 3\%–4\% improvements in average end accuracy under memory-constrained settings, with limited computational overhead.
\end{abstract}




\maketitle

\begin{figure}[t]
  \centering
  \includegraphics[width=\linewidth]{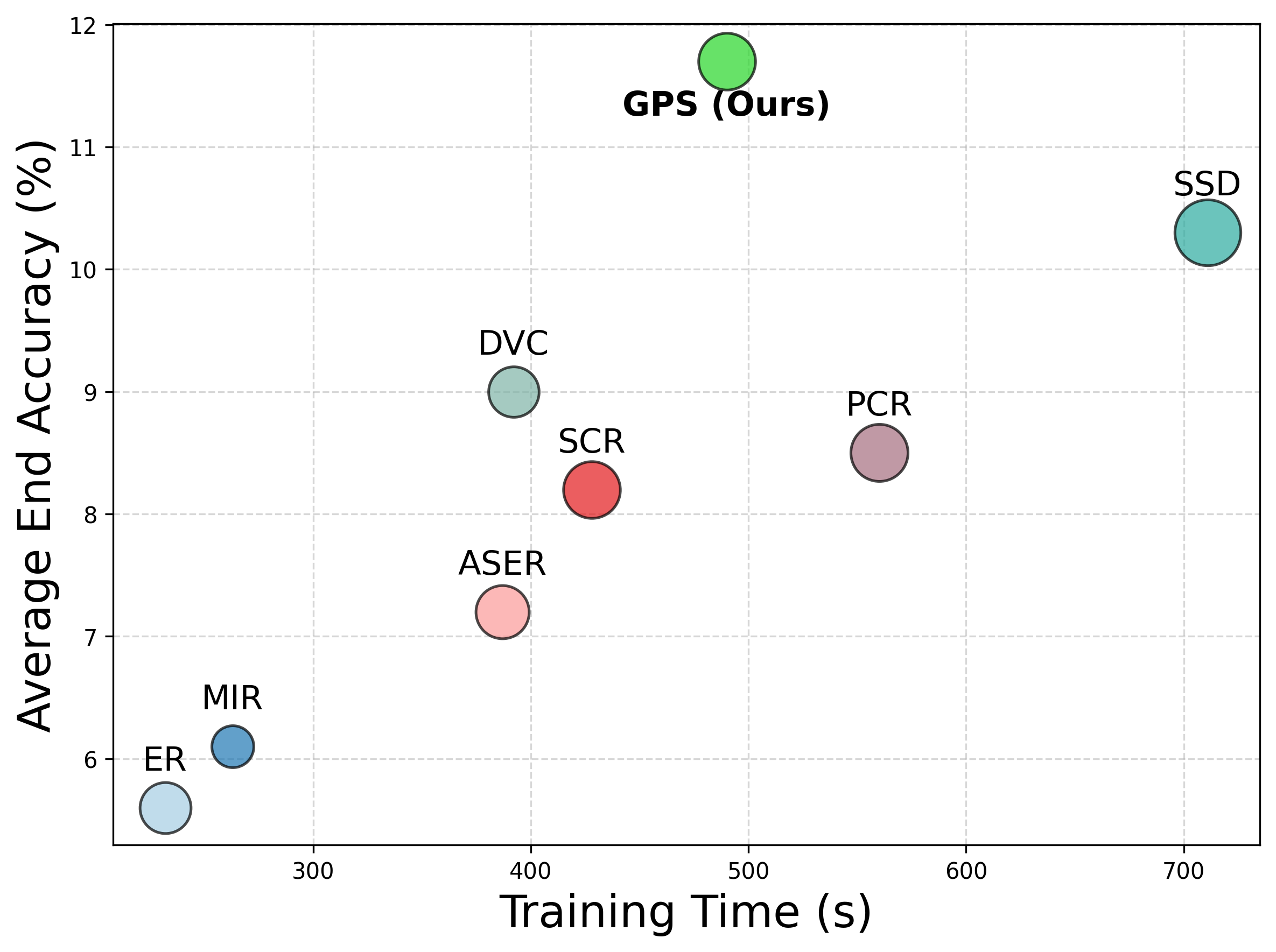}
  \caption{Efficiency vs. performance trade-off on Mini-ImageNet with a buffer size of 100. The size of each bubble reflects GPU memory usage during training. Compared to existing baselines, our method GPS achieves the highest accuracy while maintaining comparable training time.}
  \label{fig:teaser}
\end{figure}

\section{Introduction}
With the rapid growth of visual data in real-world applications, 
it is increasingly infeasible to store all past data for retraining
under the standard i.i.d. assumption. 
This challenge is further exacerbated in dynamic environments, where data distributions shift over time and storage or reuse of past samples is often limited by memory, or latency constraints.
Simply fine-tuning models on new data leads to \textit{catastrophic forgetting}~\cite{mccloskey1989catastrophic}, a 
phenomenon where previously acquired knowledge is overwritten 
by new information. 
To address this, Continual Learning (CL) ~\cite{ring1997child, thrun1998lifelong}has been 
proposed to enable models to learn from non-stationary data streams 
without access to the full history. In this work, we investigate 
the setting of online class-incremental learning, where new classes 
appear over time, data arrives sequentially, and each example is 
observed only once~\cite{chaudhry2018efficient, chaudhry2019tiny}.

To address catastrophic forgetting, existing continual 
learning methods can be categorized into three 
groups: regularization-based approaches~\cite{kirkpatrick2017overcoming, li2017learning, smith2021always, sun2023regularizing}, 
parameter-isolation strategies~\cite{mallya2018piggyback,wang2022beef, hu2023dense, xu2018reinforced}, and replay-based methods~\cite{aljundi2019online, aljundi2019gradient}. 
Among these, replay-based methods have proven especially effective in the class-incremental setting due to their simplicity and strong empirical performance~\cite{de2021continual, mai2022online}. 
These methods maintain a small memory buffer to store a subset of previously seen samples, which are replayed alongside incoming data during training. 
By revisiting past examples, the model can retain earlier knowledge while learning 
new classes, significantly reducing forgetting under constrained memory.

\begin{figure}[t]
  \centering
  \includegraphics[width=\linewidth]{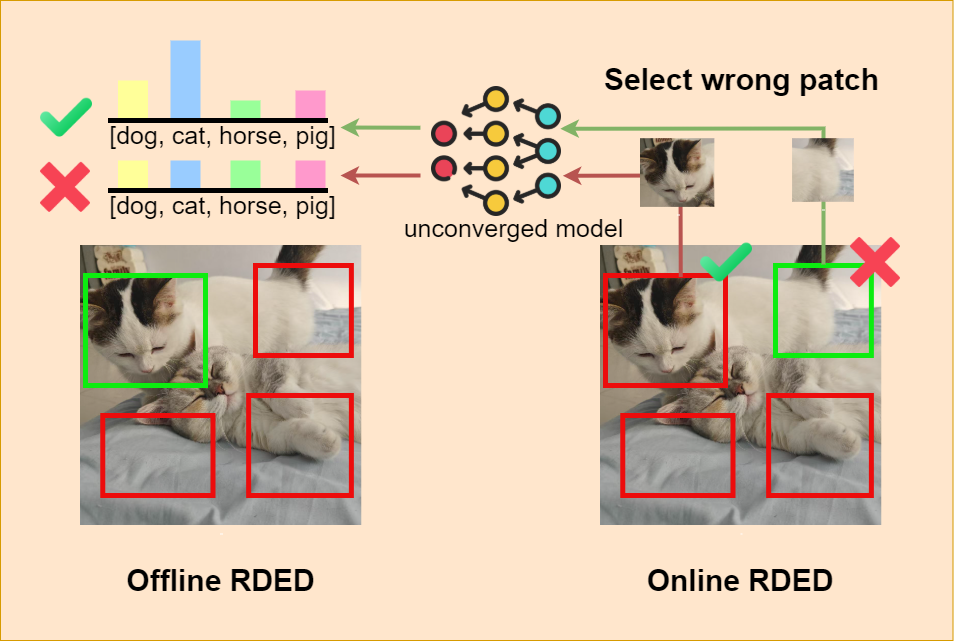}
  \caption{Illustration of RDED's limitation in the online continual learning setting. When the model is unconverged, it may select misleading or suboptimal patches due to unreliable feature representations.}
  \label{fig:rded}
\end{figure}

Recent developments in replay-based methods have explored the use of data distillation techniques to enhance memory informativeness~\cite{sangermano2022sample, gu2024summarizing}. 
However, such approaches often incur significant computational overhead due to the use of bi-level optimization ~\cite{cazenavette2022dataset, sun2024diversity}, as shown in Figure~\ref{fig:teaser}. 
To alleviate this cost, RDED ~\cite{sun2024diversity} proposes a more efficient alternative based on patch-level construction, selecting salient regions from input images without iterative optimization.
While effective in offline distillation scenarios, RDED assumes access to a well-trained feature extractor—a condition that does not hold in online continual learning, where the model remains unconverged in early stages, as illustrated in Figure~\ref{fig:rded}. 
These limitations highlight a key challenge: how to construct informative memory buffer without relying on a converged model or computationally expensive procedures.

To address this challenge, we propose a simple yet effective strategy termed Grid-based Patch Sampling (GPS) to generate informative memory samples without relying on a trainable model. 
Specifically, we partition the input image into a uniform grid of square patches. From each patch, a single pixel is sampled and aggregated to form a low-resolution image that preserves the original spatial layout.
During replay, these GPS samples are either concatenated to reconstruct an image of original resolution for training, or upsampled for use in a Nearest-Class-Mean (NCM) classifier~\cite{mensink2013distance}. 
This strategy is lightweight and model-agnostic, requiring neither bi-level optimization nor a converged feature extractor. 
By storing compact low-resolution representations, GPS allows more samples to be retained within the same memory budget, leading to increased sample diversity and higher information density in the memory buffer.

The contributions of this study can be summarized as follows:

(1) To the best of our knowledge, this is the first work to explore sampling strategy for distilling informative memory buffer in continual learning.

(2) We design a grid-based sampling mechanism that structurally compresses images into compact representations without model supervision. This enables efficient memory construction with training time comparable to standard baselines, while significantly improving memory utilization and sample capacity.

(3) Extensive experiments demonstrate that our approach 
can be seamlessly integrated into multiple continual 
learning baselines. Under memory-constrained settings, it 
improves the average end accuracy by 3\%–4\% over  
replay-based baselines.

\section{Related Work}

\subsection{Continual Learning}

Continual Learning (CL) aims to train models that can acquire and 
retain knowledge across a sequence of learning experiences. A central 
challenge in this setting is \textit{catastrophic forgetting}, where 
updating the model with new data can degrade performance on previously 
learned tasks. To address this, various approaches have been proposed, 
typically falling into three categories ~\cite{parisi2019continual} : regularization-based methods~\cite{kirkpatrick2017overcoming, li2017learning, smith2021always, sun2023regularizing}, 
which limit updates to parameters considered important for past 
tasks; parameter-isolation methods~\cite{mallya2018piggyback,wang2022beef, hu2023dense, xu2018reinforced}, which assign dedicated 
model components to different tasks; and replay-based 
methods~\cite{chaudhry2019tiny, aljundi2019online, guo2022online, mai2021supervised, lin2023pcr}, which maintain a memory buffer of past examples and use 
them during training to reinforce prior knowledge. Among these, 
replay-based approaches have shown strong performance in class-incremental 
scenarios.

\subsection{Online Continual Learning}

Online Continual Learning (OCL) addresses the challenge of learning 
from a continuous stream of data where each sample is observed only 
once and must be processed immediately. Unlike offline settings, models 
in OCL cannot store or revisit the full training set, and typically 
lack access to a well-trained backbone model. Replay-based approaches, 
particularly those derived from experience replay (ER)~\cite{chaudhry2019tiny}, 
have shown strong performance under this setting by maintaining a small 
memory buffer. Existing works in this area can be broadly divided into 
three strategies: (1) \textit{memory retrieval} methods that prioritize 
which samples to replay during training, such as MIR~\cite{aljundi2019online} and ASER~\cite{shim2021online}; 
(2) \textit{memory update} strategies that aim to select the most representative 
or informative samples for storage~\cite{aljundi2019gradient, gu2024summarizing}; and (3) \textit{model update} 
approaches that adapt the learning process to be more robust in an online 
setting~\cite{mai2021supervised, lin2023pcr, wei2023online,wang2023cba, de2021continu}. These methods collectively aim to mitigate forgetting and 
improve generalization despite the strict memory and computation constraints of OCL.

\subsection{Dataset Distillation}

Dataset distillation aims to synthesize a small yet informative set of 
training samples that can approximate the performance of a model trained 
on the full dataset~\cite{wang2018dataset}. Many existing methods formulate this as a 
bi-level optimization problem, where synthetic images are learned to 
minimize the loss on a target model trained over multiple steps.
Techniques in this line include matching training 
trajectories~\cite{cazenavette2022dataset,du2023minimizing, cui2022dc, cui2023scaling, guo2023towards, yu2023dataset}, leveraging gradient 
alignment~\cite{kim2022dataset, zhang2023accelerating, zhao2020dataset}, or incorporating regularization and 
generative priors~\cite{cazenavette2023generalizing}. While effective, these approaches 
often computationally expensive and rely on repeated model updates.
More recent work RDED~\cite{sun2024diversity} 
introduces an efficient dataset distillation paradigm. Instead of learning 
synthetic data via optimization, RDED constructs realistic samples by 
selecting high-quality patches from original images and recombining them 
into new examples. 
Although RDED improves efficiency, its depends on a converged feature extractor, making it unsuitable for online continual learning. 
These limitations underscore the need for an efficient, model-independent distillation strategy tailored to memory-constrained online settings.

\begin{figure*}[t]
  \centering
  \includegraphics[width=\textwidth]{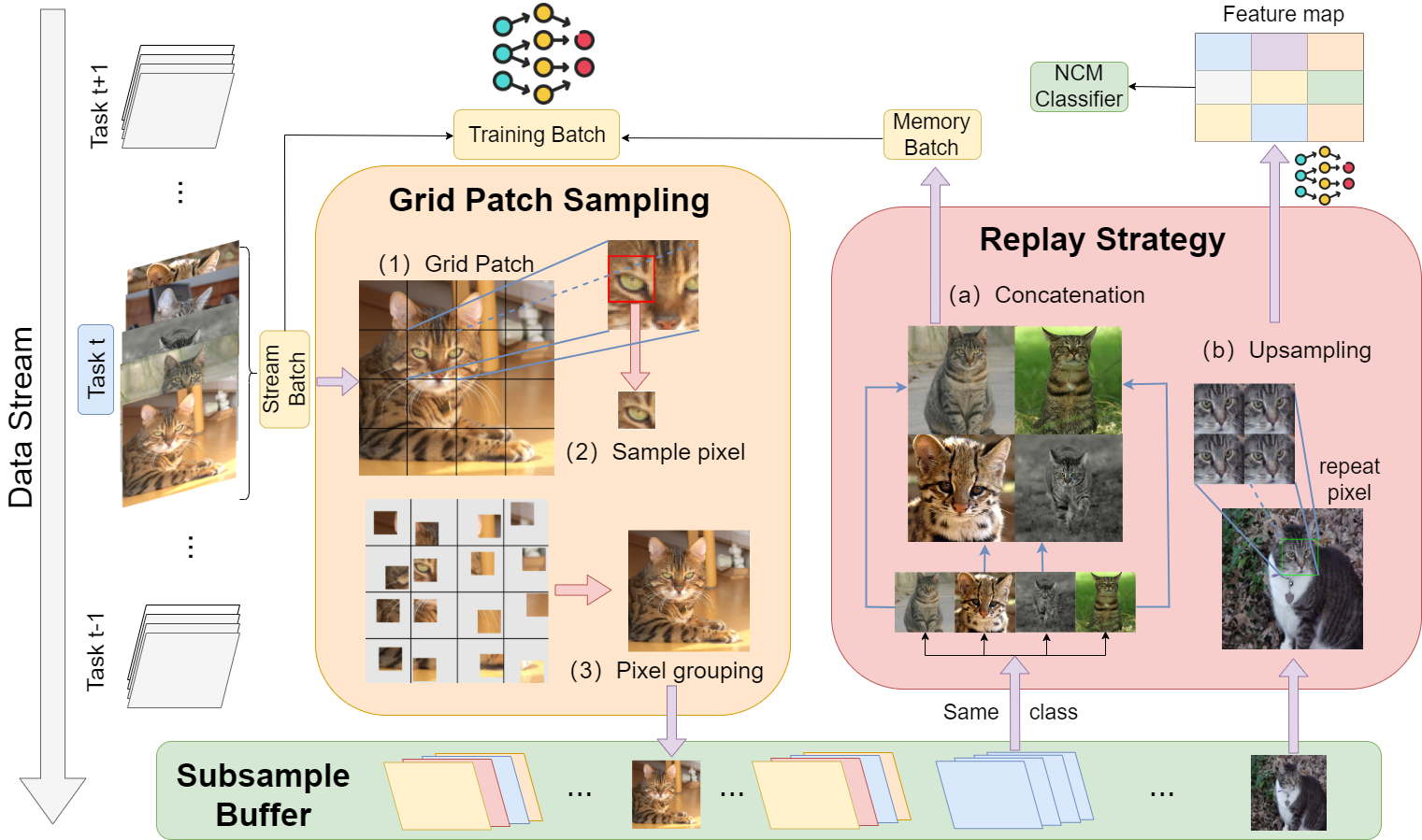}
  \caption{Overview of the proposed Grid-based Patch Sampling (GPS) framework.Given a data stream with sequential tasks, each incoming sample is processed by the GPS module: (1) the image is partitioned into a uniform grid, (2) one representative pixel is randomly sampled from each patch, and (3) sampled pixels are grouped to form a low-resolution surrogate. During replay, two strategies are used: (a) concatenation of GPS samples from the same class to reconstruct high-resolution training images; and (b) upsampling each GPS sample for inference using a Nearest Class Mean (NCM) classifier.For visualization purposes, a region is used to represent a single pixel in this image. }
  \label{fig:main}
\end{figure*}

\section{Method}

\subsection{Preliminaries}

We study the problem of online class-incremental learning ~\cite{rebuffi2017icarl, castro2018end, shmelkov2017incremental, wu2018incremental}, where a model receives a sequence of data batches $\mathcal{D} = \{\mathcal{D}_1, \mathcal{D}_2, \ldots, \mathcal{D}_T\}$, with each task $\mathcal{D}_t = \{(x_{t,i}, y_{t,i})\}_{i=1}^{n_t}$ introducing new classes not observed in previous stages. The model must learn continuously from this stream under strict constraints: each sample can be accessed only once at arrival, and no task identity is provided during inference. Class labels across tasks are mutually exclusive, i.e. $y_{t,i} \in \mathcal{C}_t$, and $\mathcal{C}_t \cap \mathcal{C}_{t'} = \emptyset$ for $t \ne t'$.

To tackle this setting, replay-based methods consider a model $\Phi$, which is updated incrementally using both current data and a compact memory buffer $\mathcal{M} = \{(x_m, y_m)\}_{m=1}^{K}$. The buffer holds a limited set of previously seen samples. During training at task $t$, the model is updated with a mini-batch drawn from the current stream $\mathcal{B}_t \subseteq \mathcal{D}_t$ and a batch from memory $\mathcal{B}_m \subseteq \mathcal{M}$. The overall training loss combines the current task loss and a replay loss:

\begin{equation}
\min_{\phi} \ \mathcal{L}_{\text{task}}(\phi; \mathcal{B}_t) + \lambda \mathcal{L}_{\text{replay}}(\phi; \mathcal{B}_m),
\end{equation}
where $\lambda$ controls the balance between acquiring new knowledge and retaining past information.

Recent work ~\cite{sangermano2022sample} has explored distilling raw samples into more informative representations to better utilize the limited capacity of the memory buffer. Methods such as SSD~\cite{gu2024summarizing} adopt data distillation to improve information density, but typically rely on bi-level optimization or relationship matching, incurring substantial computational overhead.

To address this, we introduce a lightweight alternative that directly distills semantic information from input images via spatial sampling, without requiring a trained model or costly optimization procedures. This design is particularly suited for online continual learning scenarios, where computational resources and access to well-trained backbone are both limited.

\subsection{Grid-based Patch Sampling}
To compress visual information while preserving semantically meaningful structures, we propose a simple and efficient distilling strategy termed Grid-based Patch Sampling (GPS). GPS transforms high-resolution input images into compact surrogates by uniformly sampling representative pixels from spatially localized patches. This section details the sampling mechanism and the construction of our subsample memory buffer.

\textbf{Patch Sampling Strategy.}
For simplicity, We assume that input images are square-shaped with resolution $r \times r$. Given a sampling factor $f \in \mathbb{N}$, the image is partitioned into a uniform grid of $r' \times r'$ non-overlapping square patches, where:

\begin{equation} r' = \left\lfloor \frac{r}{f} \right\rfloor. \end{equation}

Each patch covers a region of size $f \times f$ pixels. From each patch $P_{i,j}$ located at grid position $(i, j)$, we randomly select a pixel:

\begin{equation} x_{i,j} = P_{i,j}(u, v), \quad u, v \sim \mathcal{U}(0, f-1), \end{equation}
where $x_{i,j}$ is the selected pixel from patch $P_{i,j}$ and $\mathcal{U}$ denotes the uniform distribution over valid pixel indices within the patch. The set of all sampled pixels is denoted as:

\begin{equation} x^{\text{GPS}} = {x_{i,j}}_{i=1, j=1}^{r', r'}, \end{equation}
forming a compact representation image of size $r' \times r'$ that retains the spatial layout of the original input.

Overall, GPS offers an efficient and effective compression mechanism, reducing each image from $r^2$ to $(r/f)^2$ pixels with negligible computational cost. Despite the significant reduction in resolution, this sampling process preserves coarse structural information and high-level semantic cues by ensuring wide spatial coverage across the image.A qualitative visualization of this process can be found in the \hyperref[sec:visualization]{4.4} section.

\textbf{Subsample Buffer Construction.  }
To ensure a fair comparison under equal memory constraints, we quantify the memory budget in terms of total pixels. For a conventional buffer storing $K$ full-resolution images, the overall memory cost is:

\begin{equation} \text{Memory}_{\text{full}} = K \cdot r^2. \end{equation}

Under GPS, each image has only $(r/f)^2$ pixels. To maintain the same memory budget, the GPS-based buffer can store:

\begin{equation} K_{\text{GPS}} = \frac{K \cdot r^2}{(r/f)^2} = K \cdot f^2 \end{equation}
GPS-compressed images. That is, for the same memory cost, we can store $f^2$ times more images. This significantly increases the sample diversity in the buffer, which benefits generalization and knowledge retention.

To manage buffer updates under a streaming setting, we employ reservoir sampling ~\cite{chaudhry2019tiny}, which ensures uniform retention probability across all seen samples. Additionally, we maintain a class-index map that associates each class with the indices of its corresponding exemplars in the buffer, enabling efficient class-wise retrieval during replay.

By trading resolution for quantity, the GPS-based buffer retains a greater number of representative samples within the same memory budget, effectively enhancing data coverage and improving the quality of replay under memory-constrained online continual learning settings.

\subsection{Replay Strategy}
To leverage the GPS-transformed memory buffer during training and inference, two replay strategies are designed to accommodate the structure of GPS samples:  
(1) a \textit{concatenation-based strategy} for model training, and  
(2) an \textit{upsampling-based strategy} for nearest-class-mean (NCM) classification.  
These strategies enable efficient utilization of structurally compressed inputs for both parameter updates and inference.

\textbf{Concatenation-Based Replay for Training.  }
Although each GPS sample contains only coarse-grained structural information, aggregating multiple low-resolution samples of the same class allows reconstruction of training inputs that approximate the original semantic content ~\cite{sun2024diversity}. To this end, we organize GPS samples into composite images through structured concatenation.

Based on the class-index map maintained in memory, the list of image indices $\mathcal{I}_c$ corresponding to each class $c$ is first retrieved. This list is then partitioned into non-overlapping subsets of length $f^2$, where incomplete subsets are discarded. Each resulting subset defines a group of GPS samples $\{x_1, x_2, \dots, x_{f^2}\}$, all sharing label $c$.

Next, these $f^2$ low-resolution images are arranged into a $f \times f$ grid to construct a high-resolution image of size $r \times r$, denoted as:
\begin{equation}
\tilde{x}_c = \text{GridConcat}(\{x_i\}_{i=1}^{f^2}) \in \mathbb{R}^{r \times r \times C},
\end{equation}
where $C$ is the number of channels (e.g., $C=3$ for RGB images),and $x_i \in \mathbb{R}^{r' \times r' \times C}$.

The label of each reconstructed image $\tilde{x}_c$ is assigned as $y = c$, consistent with the labels of all constituent patches. This procedure yields a collection of synthetic full-resolution samples $\tilde{\mathcal{M}}$ suitable for training.

During training, a batch of $k$ reconstructed samples is randomly selected from $\tilde{\mathcal{M}}$ and combined with incoming data from the stream to update model parameters via supervised learning. This enhances sample diversity while maintaining low memory cost.

\textbf{Upsampling-Based Replay for Inference.  }To support prototype-based classification ~\cite{mai2021supervised}, the Nearest Class Mean (NCM) classifier ~\cite{mensink2013distance} is employed. For each class $c$, a mean embedding vector $\mu_c$ is computed using its associated exemplars. Given an input $x$, the classifier predicts its label by identifying the closest prototype in feature space:
\begin{equation}
\mu_c = \frac{1}{n_c} \sum_{i=1}^{n_c} f(x_i), \quad
y^* = \arg\min_{c=1,\dots,t} \| f(x) - \mu_c \|_2,
\end{equation}
where $f(\cdot)$ is the feature extractor and $n_c$ denotes the number of exemplars for class $c$.

Since GPS images have low-resolution and are incompatible with standard input size $r \times r$, each image $x \in \mathbb{R}^{r' \times r' \times C}$ is upsampled to match the input dimensionality. The upsampling is performed by repeating each pixel value $f^2$ times to reconstruct its corresponding $f \times f$ patch:
\begin{equation}
\hat{x} = \text{Upsample}(x) \in \mathbb{R}^{r \times r \times C}.
\end{equation}

These upsampled images $\hat{x}$ are then passed through the embedding network $f(\cdot)$ to compute class-wise prototypes $\mu_c$. Despite the loss in detail from sampling, the upsampled samples retain sufficient semantic information to form reliable class representations, enabling effective inference with NCM in class-incremental settings, as illustrated in Figure ~\ref{fig:tsne}.

\begin{table*}[t]
\centering
\caption{Average end accuracy (\%) on three datasets under different memory budgets ($K$). $K$ refers to the number of original images that can be stored in the buffer. Results are averaged over 10 runs.}
\label{tab:main_results}
\begin{tabular}{cc|ccc|ccc|ccc}
\toprule
\multicolumn{2}{c|}{} & \multicolumn{3}{c|}{Mini-ImageNet [84x84]} & \multicolumn{3}{c|}{CIFAR-100[32x32]} & \multicolumn{3}{c}{Tiny-ImageNet[64x64]} \\
 \multicolumn{2}{c|}{Method} & $K{=}100$ & $K{=}200$ & $K{=}500$ & $K{=}100$ & $K{=}200$ & $K{=}500$ & $K{=}200$ & $K{=}400$ & $K{=}1000$ \\
\midrule
\multicolumn{2}{c|}{fine-tune} &  & 4.2{\tiny ±0.3} &  &  & 5.8{\tiny ±0.2} &  &  & 2.3{\tiny ±0.2} &  \\
\multicolumn{2}{c|}{iid offline} &  & 47.7{\tiny ±0.2} &  &  & 50.2{\tiny ±0.3} &  &  & 25.9{\tiny ±0.4} &  \\
LwF ~\cite{li2017learning} & TPAMI 2017 & & 9.6{\tiny ±0.6} &  &  & 12.2{\tiny ±0.6} &  &  & 6.0{\tiny ±0.3} &  \\
EWC++ ~\cite{chaudhry2018riemannian} & ECCV 2018 & & 4.7{\tiny ±0.5} &  &  & 5.2{\tiny ±0.4} &  &  & 2.1{\tiny ±0.3} &  \\
\midrule
A-GEM ~\cite{chaudhry2018efficient} & ICLR 2018 & 5.0{\tiny ±0.4} & 5.2{\tiny ±0.4} & 5.0{\tiny ±0.5} & 5.7{\tiny ±0.4} & 5.9{\tiny ±0.4} & 5.9{\tiny ±0.3} & 2.8{\tiny ±0.2} & 3.0{\tiny ±0.2} & 2.6{\tiny ±0.3} \\
ER ~\cite{chaudhry2019tiny} & arXiv 2019 & 5.6{\tiny ±0.6} & 6.0{\tiny ±0.4} & 7.1{\tiny ±1.1} & 7.2{\tiny ±0.4} & 7.4{\tiny ±0.5} & 9.4{\tiny ±0.6} & 3.2{\tiny ±0.2} & 3.2{\tiny ±0.3} & 3.6{\tiny ±0.4} \\
MIR ~\cite{aljundi2019online} & NeurIPS 2019 & 6.1{\tiny ±0.5} & 6.7{\tiny ±0.3} & 8.6{\tiny ±1.1} & 7.3{\tiny ±0.5} & 7.6{\tiny ±0.4} & 10.0{\tiny ±0.5} & 3.6{\tiny ±0.2} & 3.9{\tiny ±0.3} & 4.6{\tiny ±0.6} \\
ASER$_\mu$ ~\cite{shim2021online} & AAAI 2021 & 7.2{\tiny ±0.4} & 8.8{\tiny ±0.5} & 12.0{\tiny ±0.8} & 7.8{\tiny ±0.2} & 8.4{\tiny ±0.5} & 11.0{\tiny ±0.4} & 3.7{\tiny ±0.2} & 5.8{\tiny ±0.3} & 8.4{\tiny ±0.7} \\
SCR ~\cite{mai2021supervised} & CVPR 2021 & 8.2{\tiny ±0.4} & 11.3{\tiny ±0.5} & 18.3{\tiny ±0.7} & 9.1{\tiny ±0.4} & 12.1{\tiny ±0.7} & 19.7{\tiny ±0.4} & 4.8{\tiny ±0.3} & 6.9{\tiny ±0.4} & \textbf{12.1{\tiny ±0.4}} \\
DVC ~\cite{gu2022not} & CVPR 2022 & 9.0{\tiny ±0.4} & 10.3{\tiny ±0.9} & 14.9{\tiny ±1.3} & 10.6{\tiny ±0.6} & 13.6{\tiny ±0.7} & 19.1{\tiny ±0.9} & 6.1{\tiny ±0.4} & 7.4{\tiny ±0.7} & 9.6{\tiny ±0.9} \\
PCR ~\cite{lin2023pcr} & CVPR 2023 & 8.5{\tiny ±1.1} & 10.0{\tiny ±1.1} & 14.5{\tiny ±1.1} & 10.6{\tiny ±1.0} & 13.3{\tiny ±0.6} & 19.6{\tiny ±0.6} & 6.0{\tiny ±0.5} & 7.2{\tiny ±0.8} & 9.6{\tiny ±0.5} \\
SSD ~\cite{gu2024summarizing} & AAAI 2024 & 10.3{\tiny ±0.6} & 13.4{\tiny ±0.6} & 19.8{\tiny ±0.4} & 9.5{\tiny ±0.4} & 13.7{\tiny ±0.5} & \textbf{22.2{\tiny ±0.6}} & 3.3{\tiny ±0.2} & 4.6{\tiny ±0.2} & 6.2{\tiny ±0.2} \\
GPS (Ours) & This paper & \textbf{11.7{\tiny ±0.5}} & \textbf{15.4{\tiny ±0.7}} & \textbf{21.2{\tiny ±0.7}} & \textbf{11.2{\tiny ±1.0}} & \textbf{15.4{\tiny ±0.9}} & 18.7{\tiny ±1.3} & \textbf{6.5{\tiny ±0.4}} & \textbf{8.9{\tiny ±0.5}} & 11.4{\tiny ±1.7} \\
\bottomrule
\end{tabular}
\end{table*}

\subsection{Efficiency and Practical Consideration}

\textbf{Computational Simplicity.  }
GPS is designed to be lightweight and efficient, requiring no additional training procedures or optimization steps. Unlike previous distillation-based methods that depend on complex bi-level optimization or auxiliary networks, GPS performs a simple patch sampling process directly on the input image. This process is non-parametric and operates independently of any model predictions or gradient signals, making it entirely model-free. Moreover, it does not rely on any pre-trained backbone. As a result, GPS introduces minimal computational overhead and is well-suited for real-time or resource-constrained continual learning scenarios.

\textbf{Memory Efficiency.  } 
By reducing each image from $r^2$ to $(r/f)^2$ pixels, GPS achieves a compression ratio of $f^2$, enabling the storage of $f^2$ times more samples within the same memory budget. This expansion in buffer capacity significantly increases sample diversity and allows better coverage of the feature space across classes. Despite the aggressive downsampling, the retained structural information enables effective learning and classification, particularly when combined with concatenation-based replay and prototype-based inference.

\textbf{Plug-and-Play Integration.} 
GPS operates entirely at the data level, requiring no changes to model architecture or training objectives. This design makes it readily compatible with a wide range of replay-based continual learning methods, allowing seamless integration into existing pipelines to enhance memory efficiency. Its simplicity and generality also make it particularly well-suited for deployment in real-world scenarios where model internals are fixed or computational resources are limited.

\section{EXPERIMENTS}

\subsection{Experiment Settings}

\textbf{Datasets. } We evaluate our method on three widely used benchmarks in online class-incremental learning: Sequential CIFAR-100 ~\cite{krizhevsky2009learning}, Sequential MiniImageNet ~\cite{vinyals2016matching}, and Sequential TinyImageNet ~\cite{deng2009imagenet}. Following previous works ~\cite{vinyals2016matching}, CIFAR-100 is split into 10 tasks with 10 classes per task; Mini-ImageNet follows the same partitioning scheme with 10 tasks of 10 classes each, and Tiny-ImageNet is divided into 20 tasks, each consisting of 10 unique classes.

\textbf{Evaluation Metric. } 
We adopt the standard \textit{average end accuracy} as our evaluation metric ~\cite{chaudhry2018riemannian, chaudhry2019tiny}. It measures the average test accuracy across all seen tasks after the final task has been learned. Formally, let $a_{N,i}$ denote the test accuracy on task $i$ after the model has been trained on all $N$ tasks. The average end accuracy $A_N$ is computed as:

\begin{equation}
\text{Average End Accuracy}(A_N) = \frac{1}{N} \sum_{i=1}^{N} a_{N,i}
\label{eq:avg_accuracy}
\end{equation}

To ensure statistical reliability, each experiment is repeated 10 times, and we report both the mean and standard deviation across runs.

\textbf{Baselines. }
We compare our method with representative baselines from three categories:
Non-continual learning methods: IID Offline, Fine-tune;
Regularization-based methods: LwF ~\cite{li2017learning}, EWC++ ~\cite{chaudhry2018riemannian};
Replay-based methods: A-GEM ~\cite{chaudhry2018efficient}, ER ~\cite{chaudhry2019tiny}, MIR ~\cite{aljundi2019online}, ASER ~\cite{shim2021online}, SCR ~\cite{mai2021supervised}, DVC ~\cite{gu2022not}, PCR ~\cite{lin2023pcr}, SSD ~\cite{gu2024summarizing}.

\textbf{Implementation Details. }
Following prior work~\cite{aljundi2019online, mai2021supervised, lopez2017gradient}, we adopt a reduced ResNet-18~\cite{he2016deep} as the backbone network across all datasets. Training is performed using stochastic gradient descent with a learning rate of 0.1. At each time step, the model receives a mini-batch of 10 samples from the data stream, and 100 samples are retrieved from the memory buffer for replay.
To reflect memory-constrained settings, we vary the buffer size to hold roughly 1, 2, or 5 exemplars per class.

Experience-replay-based methods are optimized with the cross-entropy loss and employ a Softmax classifier for prediction. In contrast, methods based on supervised contrastive replay (SCR) ~\cite{gu2024summarizing} are trained with the supervised contrastive loss (SCL) ~\cite{khosla2020supervised} and utilize the NCM classifier during inference.
Our method is implemented on top of the SCR framework. The sampling factor for Grid-based Patch Sampling (GPS) is set to $f=2$ across all experiments. For replay: during training, retrieved GPS samples are processed using the concatenation-based reconstruction strategy; during inference with the NCM classifier, GPS samples are upsampled to match the input resolution.

\subsection{Comparison with SOTA}

\textbf{Comparison of Average End Accuracy. }
As shown in Table~\ref{tab:main_results}, our method (GPS) consistently achieves the highest average end accuracy across most evaluated settings. This demonstrates the effectiveness of our structure-preserving sampling strategy in increasing the information density of the memory buffer under constrained capacity.

In particular, GPS outperforms SSD, a recent data distillation approach designed to improve memory efficiency through synthetic sample construction. While SSD enhances buffer quality, it relies on computationally intensive optimization procedures. In contrast, GPS achieves better performance without such overhead, highlighting its advantages in both effectiveness and efficiency. A detailed comparison of computational cost is provided in the next section.

Notably, the benefits of GPS become even more pronounced under low-memory settings. When the buffer size is limited to $K{=}100$, GPS improves upon its base method SCR by 3.5\%. This indicates its ability to make the most of restricted memory through spatially structured sampling and increased exemplar-level diversity.

On the lower-resolution CIFAR-100 dataset, GPS achieves the best performance when memory is scarce ($K{=}100$), but is slightly outperformed by SCR as memory increases. We attribute this to the fact that, at low resolutions, the gain from storing more compressed samples is no longer sufficient to compensate for the degradation in semantic fidelity. Nevertheless, considering that high-resolution inputs are more common in real-world scenarios, the strong performance of GPS on higher-resolution datasets underscores its competitiveness in realistic continual learning applications.

\begin{figure}[h]
    \centering
    \begin{subfigure}[b]{\columnwidth}
        \includegraphics[width=\linewidth]{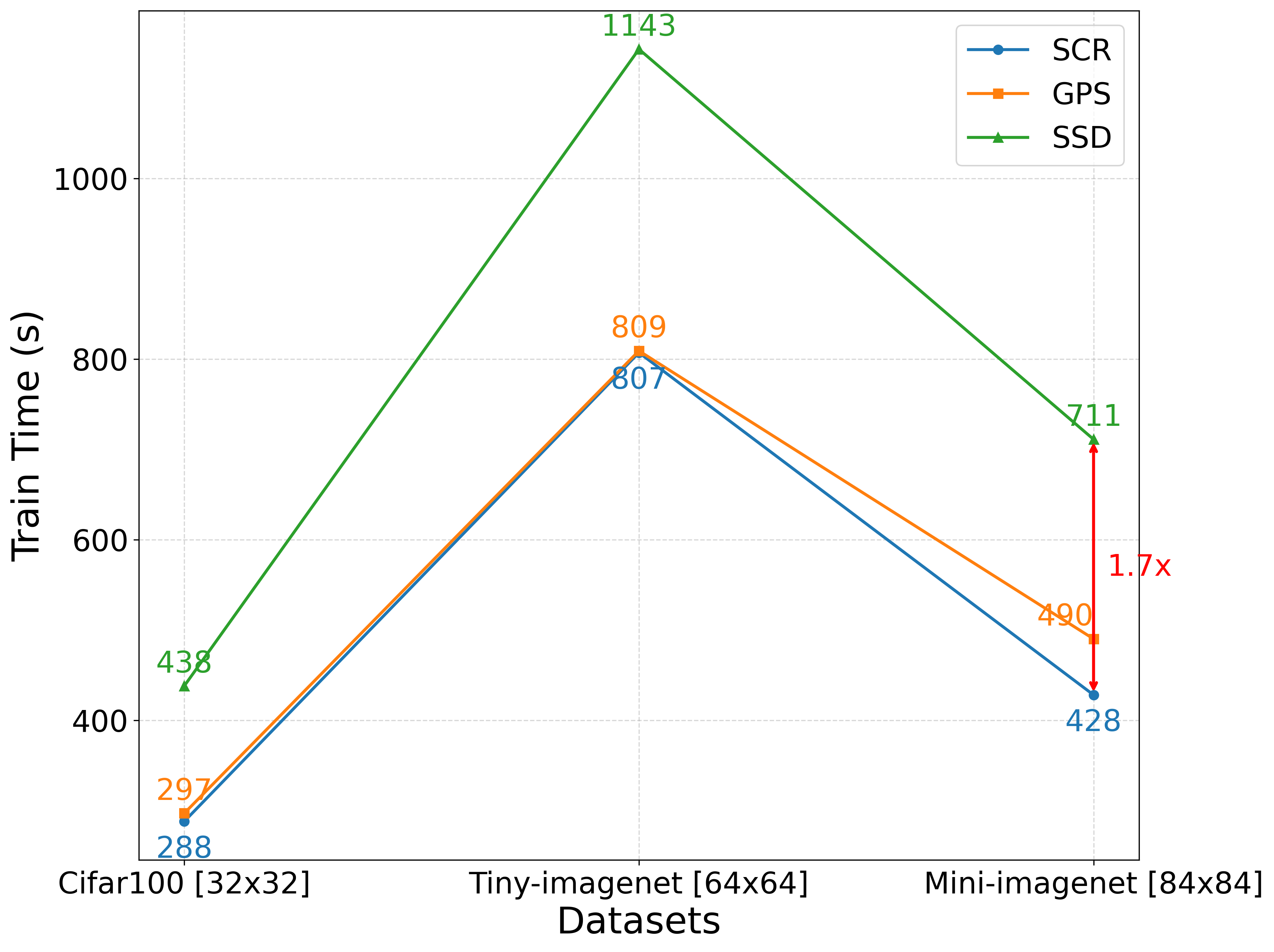}
        \caption{Training time}
        \label{fig:cost_time}
    \end{subfigure}
    \hfill
    \begin{subfigure}[b]{\columnwidth}
        \includegraphics[width=\linewidth]{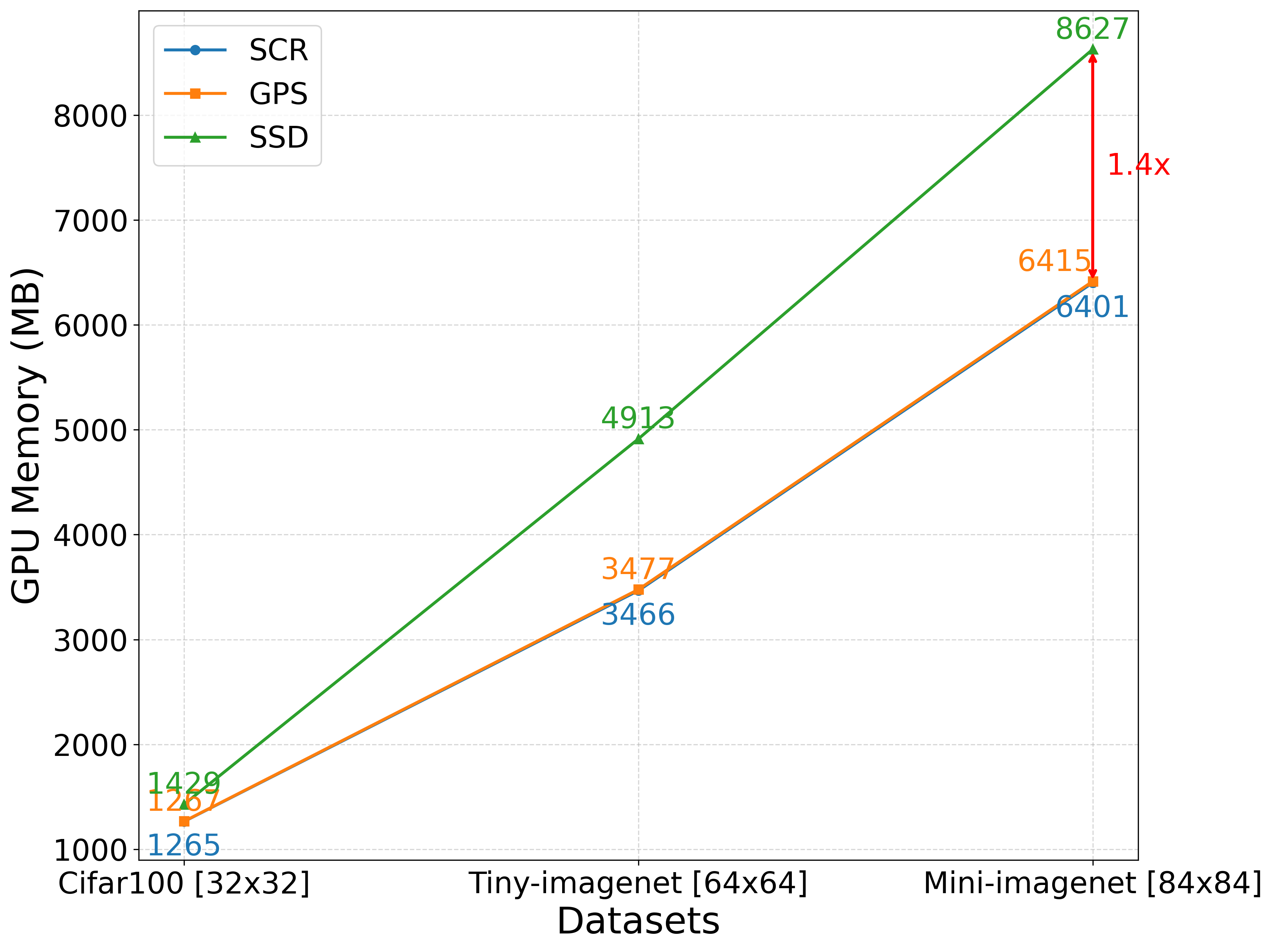}
        \caption{GPU Memory}
        \label{fig:cost_gpu}
    \end{subfigure}
    \caption{Comparison of training time (top) and GPU memory usage (down) on three datasets. We use a buffer size configured to store approximately one exemplar per class as a case study.}
    \label{fig:computation_cost}
\end{figure}

\textbf{Comparison of Computational Cost. }
Computational efficiency is critical in online continual learning, especially under real-time or resource-constrained conditions. We compare the training time and GPU memory usage of our method (GPS) with two strong baselines: SCR and SSD. Results are presented in Figure~\ref{fig:computation_cost}.

As shown in Figure~\ref{fig:cost_time}, GPS maintains training time on par with SCR across all datasets, despite processing a substantially larger number of compressed samples. In contrast, SSD incurs a substantial computational overhead, especially on high-resolution datasets. For instance, on MiniImageNet, GPS reduces training time by over 29\% compared to SSD (809s vs. 1143s), while remaining comparable to SCR (807s).

Figure~\ref{fig:cost_gpu} illustrates GPU memory usage during training. GPS and SCR exhibit nearly nearly memory footprints across datasets, whereas SSD consistently consumes more memory due to its bi-level optimization and extra storage for relationship matching. On MiniImageNet, GPS uses only 6415MB of GPU memory, compared to 8627MB for SSD—a 1.4$\times$ increase.

These results confirm that GPS achieves state-of-the-art accuracy while maintaining low computational cost. Unlike SSD, which improves performance at the expense of training time and memory usage, GPS offers a lightweight and efficient alternative. Its simplicity makes it ideal for deployment in real-world scenarios with limited hardware resources, such as mobile robotics or edge devices.

\subsection{Ablation Study}

\begin{table}
  \caption{Impact of the sampling factor $f$ on average end accuracy on Mini-ImageNet under varying buffer sizes. When $f=1$, the model stores full-resolution images and corresponds to the SCR baseline. Setting $f=2$ yields the best performance across all settings and is used as the default in our method. }
  \label{tab:f_impact}
  \begin{tabular}{c|ccc}
    \toprule
     &\multicolumn{3}{c}{Mini-ImageNet}\\
     $f$ & $K{=}100$ & $K{=}200$ & $K{=}500$ \\
    \midrule
    1 (SCR) & 8.2{\tiny ±0.4} & 11.3{\tiny ±0.5} & 18.3{\tiny ±0.7} \\
    2 (Ours) & \textbf{11.7{\tiny ±0.5}} & \textbf{15.4{\tiny ±0.7}} & \textbf{21.2{\tiny ±0.7}} \\
    4 & 5.6{\tiny ±0.5} & 5.4{\tiny ±0.2} & 4.8{\tiny ±0.7} \\
    7 & 3.5{\tiny ±0.7} & 3.7{\tiny ±1.2} & 3.8{\tiny ±0.5} \\
  \bottomrule
\end{tabular}
\end{table}

\textbf{Impact of Sampling Factor $f$.}  
We investigate the effect of the sampling factor $f$ in Grid-based Patch Sampling (GPS), which controls the spatial granularity of sampled patches. Table~\ref{tab:f_impact} reports the average end accuracy on Mini-ImageNet under different memory budgets ($K=100, 200, 500$) with varying values of $f$.Setting $f=1$ corresponds to storing full-resolution images. 
As shown in the results, $f=2$ consistently yields the best performance across all memory budgets, significantly outperforming both smaller ($f=1$) and larger ($f=4,7$) configurations. This suggests that $f=2$ provides a favorable trade-off between resolution and sample diversity. In particular, increasing $f$ beyond 2 leads to excessive downsampling, which degrades semantic fidelity and weakens the model’s ability to preserve class-discriminative features, resulting in sharp accuracy drops. 
See Figure~\ref{fig:vis_f} for visual comparison.

\begin{table}
  \caption{ Effect of applying GPS as a memory construction strategy to different baselines on Mini-ImageNet under varying buffer sizes. Across all methods and memory budgets, GPS consistently improves performance, demonstrating its generality and effectiveness in enhancing memory informativeness.}
  \label{tab:apply_baseline}
  \begin{tabular}{c|ccc}
    \toprule
     &\multicolumn{3}{c}{Mini-ImageNet}\\
     Method & $K{=}100$ & $K{=}200$ & $K{=}500$ \\
    \midrule
    ER & 5.6 & 6.0 & 7.1 \\
    ER+GPS & \textbf{7.3}(+1.7) & \textbf{6.5}(+0.5) & \textbf{7.2}(+0.1) \\
    \midrule
    PCR & 8.5 & 10.0 & 14.5 \\
    PCR+GPS & \textbf{10.1}(+1.6) & \textbf{12.1}(+2.1) & \textbf{16.1}(+1.6) \\
    \midrule
    SCR & 8.2 & 11.3 & 18.3 \\
    SCR+GPS & \textbf{11.7}(+3.5) & \textbf{15.4}(+4.1) & \textbf{21.2}(+2.9) \\
  \bottomrule
\end{tabular}
\end{table}

\textbf{Application to Different Baselines.}  
To evaluate the generality of our method, we apply GPS to three representative replay-based continual learning baselines: ER, PCR, and SCR. As shown in Table~\ref{tab:apply_baseline}, GPS consistently improves performance across all three frameworks, demonstrating its compatibility with diverse training paradigms. The gains are particularly pronounced when combined with stronger strategies such as SCR and PCR, suggesting that more effective training schemes can further amplify the benefits of structure-preserving memory sampling.
This highlights the potential of GPS as a general-purpose, plug-and-play component for enhancing memory efficiency in a wide range of continual learning systems.

\begin{table}
  \caption{Comparison of different memory construction methods applied within the same training framework (SCR) on Mini-ImageNet under varying buffer sizes ($K=100$, $200$, and $500$). }
  \label{tab:scr+}
  \begin{tabular}{c|ccc}
    \toprule
     &\multicolumn{3}{c}{Mini-ImageNet}\\
     Method & $K{=}100$ & $K{=}200$ & $K{=}500$ \\
    \midrule
    SCR & 8.2 & 11.3 & 18.3 \\
    SCR+ASER & 8.6(+0.4) & 12.2(+0.9) & 18.1(-0.2) \\
    SCR+SSD & 10.3(+2.1) & 13.4(+2.1) & 19.8(+1.5) \\
    SCR+GPS(Ours) & \textbf{11.7}(+3.5) & \textbf{15.4}(+4.1) & \textbf{21.2}(+2.9) \\
  \bottomrule
\end{tabular}
\end{table}

\textbf{Comparing with Other Methods.}  
As shown in Table~\ref{tab:scr+}, we compare GPS with other representative memory-construction strategies, including ASER and SSD, under a shared SCR training framework. While ASER offers slight improvements over the SCR baseline, its gains are marginal and diminish as memory increases, SSD achieves more substantial improvements. However, GPS surpasses both, delivering the highest accuracy across all configurations. For instance, at $K{=}200$, GPS improves end accuracy by +4.1\% over SCR, compared to +2.1\% by SSD and +0.9\% by ASER.These results demonstrate that GPS not only enhances the informativeness of stored samples more effectively than existing methods, but also maintains superior scalability and robustness across different memory size.

\begin{figure}[h]
  \centering
  \includegraphics[width=\linewidth]{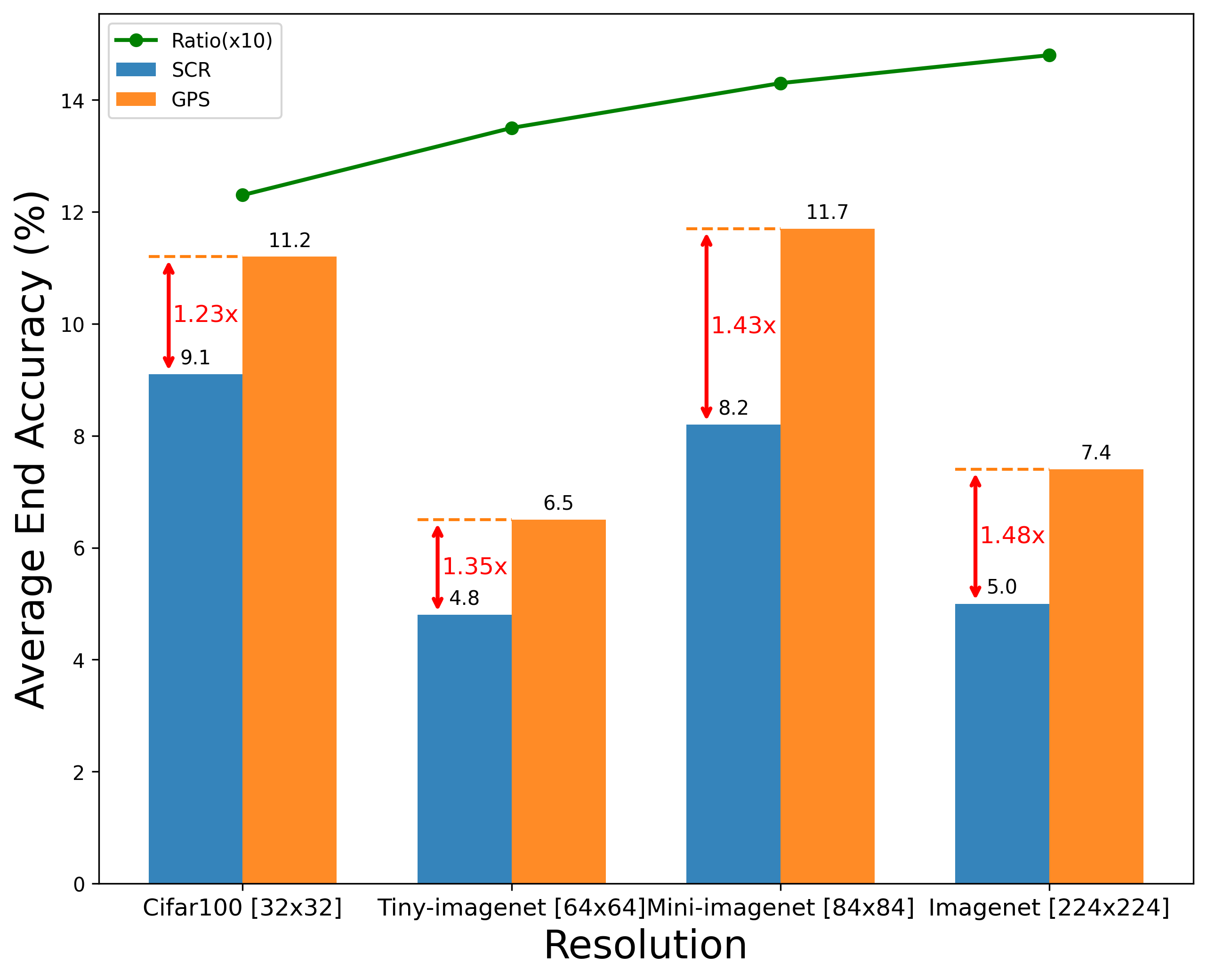}
  \caption{Average end accuracy comparison between SCR and GPS under increasing input resolutions. Experiments are conducted under a constrained buffer budget with one exemplar per class. GPS achieves larger gains at higher resolutions, demonstrating better scalability.}
  \label{fig:resolution}
\end{figure}

\textbf{Impact of image resolution.}
Beyond lower-resolution datasets, we extend our analysis to evaluate the effect of input resolution on GPS. To this end, we construct Sequential ImageNet-100 by selecting 100 classes from ImageNet ~\cite{deng2009imagenet} and splitting them into 10 tasks. Under a memory budget of one exemplar per class, we benchmark GPS against SCR across CIFAR-100 (32$\times$32), Tiny-ImageNet (64$\times$64), Mini-ImageNet (84$\times$84), and ImageNet-100 (224$\times$224).

As shown in Figure~\ref{fig:resolution}, GPS consistently outperforms SCR across all evaluated resolutions, from CIFAR-100 (32$\times$32) to ImageNet-100 (224$\times$224). Notably, the relative improvement in average end accuracy increases as the input resolution becomes higher, reaching up to a 1.48$\times$ gain on ImageNet-100. 
This trend highlights that GPS scales effectively with input resolution, as higher-resolution images provide more spatial detail for structured sampling.

We hypothesize that this ratio gain stems from the nature of our sampling mechanism: at higher resolutions, the spatial granularity of the input allows GPS to capture a more complete and informative representation of visual content, even after aggressive compression. The subsequent upsampling or concatenation processes are thus able to reconstruct semantically meaningful patterns more effectively, leading to improved classification performance. These results demonstrate that GPS is particularly well-suited for high-resolution continual learning scenarios, where the structural information preserved during sampling is more complete.

\subsection{Visualization}
\label{sec:visualization}

\begin{figure}[h]
  \centering
  \includegraphics[width=\linewidth]{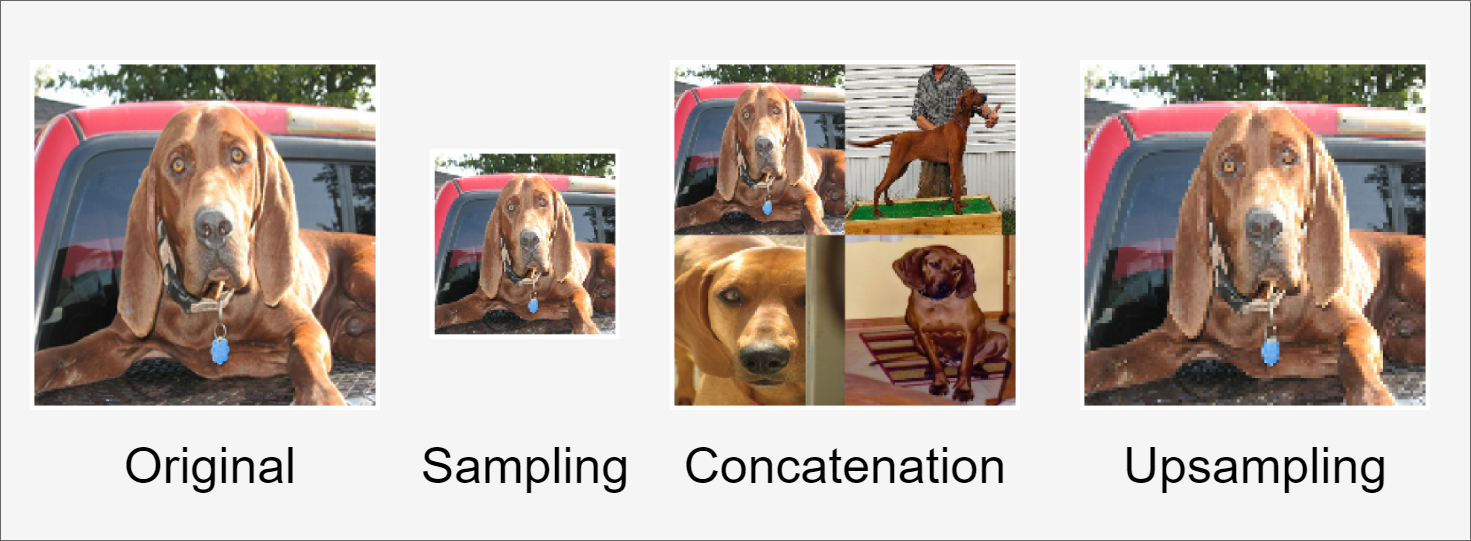}
  \caption{Visualization on Mini-imagenet. From left to right, the figure presents: the original input image, its sampled version using GPS, a reconstructed high-resolution image via patch concatenation, and the result of upsampling a single GPS sample.}
  \label{fig:visualization}
\end{figure}

\begin{figure}[h]
  \centering
  \includegraphics[width=\linewidth]{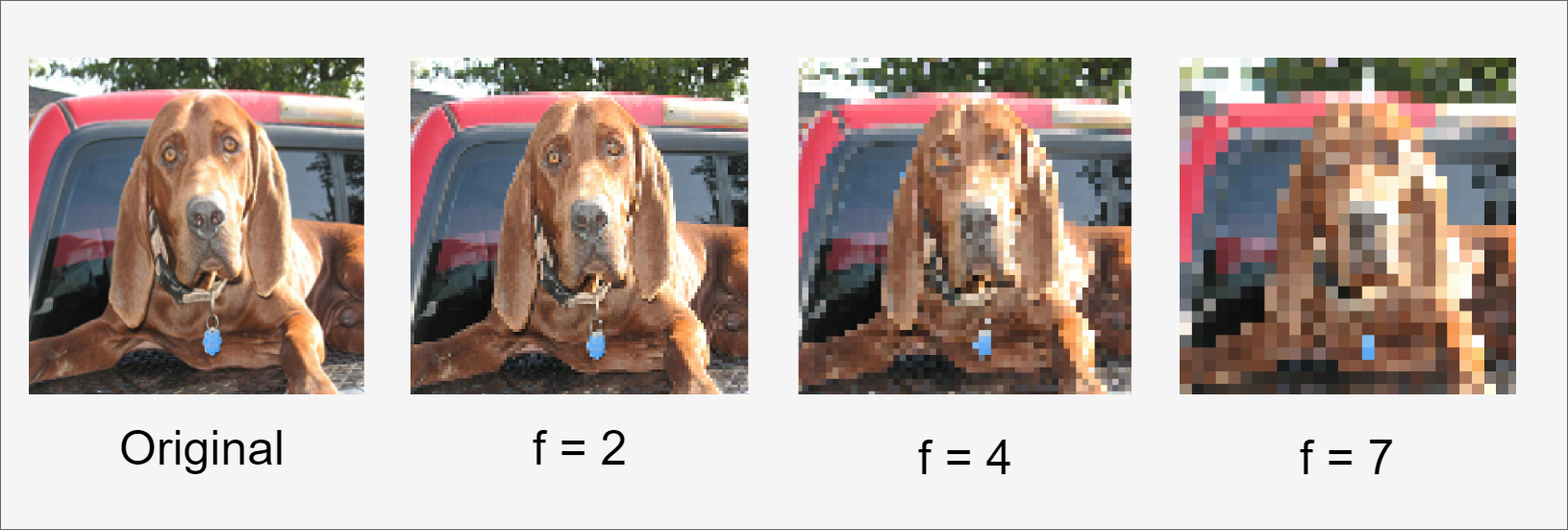}
  \caption{Visualization of GPS-processed images under different sampling factors $f$. A larger $f$ corresponds to a more aggressive reduction in resolution. While $f=2$ preserves most structural and semantic details, further increasing $f$ (e.g., to 4 or 7) leads to significant information loss and perceptual degradation. }
  \label{fig:vis_f}
\end{figure}

\textbf{Image-wise.}  
As a qualitative complement to our quantitative results, we provide visualizations of key stages in the GPS pipeline. As shown in Figure~\ref{fig:visualization}, the sampled image (via $f=2$) retains essential semantic cues such as object shape and spatial layout, despite the significant reduction in resolution. When multiple low-resolution samples are concatenated, the resulting composite image preserves class-discriminative features sufficient for training. Similarly, upsampled images, though coarser, remain structurally coherent and support reliable inference through NCM classification.

In addition, Figure~\ref{fig:vis_f} shows the effect of varying the sampling factor $f$. While $f=2$ provides a favorable trade-off between fidelity and compression, increasing $f$ further leads to substantial visual degradation, with high-frequency details and object boundaries becoming indistinct. This supports our quantitative findings that overly aggressive sampling impairs performance by reducing the discriminative capacity of stored exemplars.

\begin{figure}[h]
    \centering
    \begin{subfigure}[b]{0.48\columnwidth}
        \includegraphics[width=\linewidth]{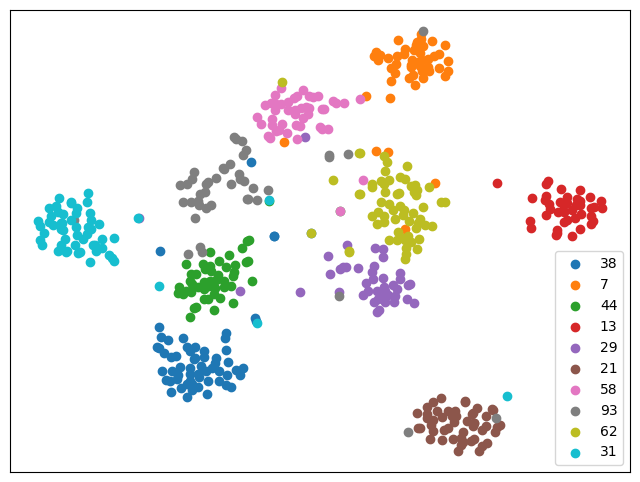}
        \caption{SCR}
        \label{fig:scr_tsne}
    \end{subfigure}
    \hfill
    \begin{subfigure}[b]{0.48\columnwidth}
        \includegraphics[width=\linewidth]{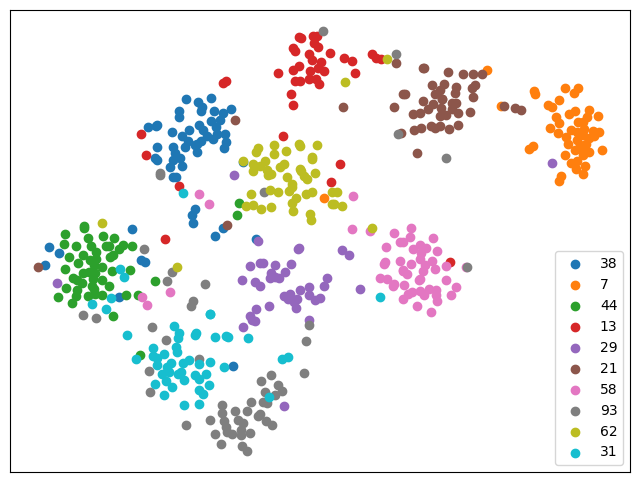}
        \caption{GPS(Ours)}
        \label{fig:gps_tsne}
    \end{subfigure}
    \caption{2D t-SNE visualization of feature embeddings from buffered samples on Mini-ImageNet after training. We randomly select 10 classes for visualization. Each point corresponds to a buffer example encoded by the feature extractor. (a) SCR baseline. (b) Our method (GPS). GPS produces more dispersed intra-class features, indicating increased sample diversity, while maintaining clear inter-class boundaries.}
    \label{fig:tsne}
\end{figure}

\textbf{Feature-wise.}  
To better understand the impact of our method on representation learning, we visualize t-SNE projections ~\cite{van2008visualizing} of feature embeddings from buffered samples on Mini-ImageNet, as shown in Figure~\ref{fig:tsne}. Compared to SCR, GPS exhibits greater intra-class dispersion for certain categories, which reflects the increased sample diversity afforded by storing a larger number of structurally distinct exemplars. While this dispersion may be partially attributed to feature shift caused by the sample-upsample pipeline, GPS still maintains clear inter-class boundaries, indicating that semantic separability is preserved. These results demonstrate that GPS enhances buffer diversity without sacrificing global feature discriminability.

\section{Conclusion}

In this work, we introduced Grid-based Patch Sampling (GPS), a novel and efficient memory construction strategy for online class-incremental learning. Unlike prior distillation-based methods that rely on heavy optimization or pre-trained models, GPS leverages a simple grid-wise sampling mechanism to generate compact, structure-preserving image surrogates. By maintaining spatial coherence and enabling the storage of significantly more exemplars within a fixed memory budget, GPS enhances both memory efficiency and sample diversity. Through extensive experiments on multiple benchmarks, we demonstrated that GPS not only achieves state-of-the-art performance under strict memory constraints, but also significantly reduces computational overhead. 
We believe this study opens up new possibilities for efficient replay in continual learning and encourages further exploration into lightweight, informative memory construction techniques.

\begin{acks}
To Robert, for the bagels and explaining CMYK and color spaces.
\end{acks}

\bibliographystyle{ACM-Reference-Format}
\bibliography{sample-base}










\end{document}